%% file: main.tex
\documentclass[10pt,twocolumn,letterpaper]{article}

\usepackage[pagenumbers]{cvpr} 

\usepackage{graphicx}
\usepackage{caption}
\usepackage{algorithm}
\usepackage{algorithmic}
\usepackage{multirow}
\usepackage{amsfonts,amsmath,amssymb}
\usepackage{enumitem}
\usepackage{pifont}
\usepackage{newfloat}
\usepackage{listings}
\usepackage{soul}
\usepackage{tikz}
\usepackage{arydshln}
\usepackage{tabularx}
\usepackage{booktabs}
\usepackage{tcolorbox}
\usepackage{wrapfig}
\usepackage{minitoc}
\usepackage{colortbl}   
\usepackage{xcolor}     

\usepackage{color}

\definecolor{cvprblue}{rgb}{0.21,0.49,0.74}
\usepackage[pagebackref,breaklinks,colorlinks,allcolors=cvprblue]{hyperref}

\title{Harmonious Parameter Adaptation in Continual Visual Instruction Tuning \\ for Safety-Aligned MLLMs}

\author{Ziqi Wang\textsuperscript{1}, Chang Che\textsuperscript{1}, Qi Wang\textsuperscript{2}, Hui Ma\textsuperscript{1}, Zenglin Shi\textsuperscript{1}\thanks{Corresponding author: zenglin.shi@hfut.edu.cn}, Cees G. M. Snoek\textsuperscript{3}, Meng Wang\textsuperscript{1}\\
{\textsuperscript{1}Hefei University of Technology }
{\textsuperscript{2}Tsinghua University }
{\textsuperscript{3}University of Amsterdam }
}

\begin{document}
\maketitle
\input{sec/0_abstract}    
\input{sec/1_intro}

\input{sec/2_related}
\input{sec/3_method}
\input{sec/4_exp}

\input{sec/5_conclusion}

{
    \small
    \bibliographystyle{ieeenat_fullname}
    \bibliography{main}
}
\input{sec/X_suppl}

\end{document}

%% file: sec/0_abstract.tex
\begin{abstract}
While continual visual instruction tuning (CVIT) has shown promise in adapting multimodal large language models (MLLMs), existing studies predominantly focus on models without safety alignment. This critical oversight ignores the fact that real-world MLLMs inherently require such mechanisms to mitigate potential risks. In this work, we shift our focus to CVIT for safety-aligned MLLMs and observe that during continual adaptation, the model not only suffers from task forgetting but also exhibits degradation in its safety. Achieving a harmonious balance between safety and task performance remains a crucial challenge. To address this, we propose Harmonious Parameter Adaptation (HPA), a post-training framework composed of focusing-based parameter partition,  harmoniously balanced parameter selection, and orthogonal parameter adjustment. Specifically, HPA partitions parameters into two types based on their focus on safety or task performance, and selects the focused ones to preserve from a balanced perspective. In addition, HPA imposes orthogonality constraints on parameter updates to further alleviate catastrophic forgetting. Extensive experiments on the CVIT benchmark and safety evaluation datasets demonstrate that HPA better maintains high safety and mitigates forgetting than existing baselines. Code is available at https://github.com/Minato-Zackie/HPA. 

\end{abstract}

%% file: sec/1_intro.tex
\section{Introduction}
\label{sec:intro}

Multimodal large language models (MLLMs) \cite{liu2024visual, wang2024qwen2}, built upon large language models (LLMs) \cite{touvron2023llama, naveed2025comprehensive}, have recently attracted significant attention for their strong vision-language reasoning capabilities. Their impressive performance stems from a two-stage training paradigm \cite{zhu2024minigpt, liu2024visual}: a pre-training stage that equips the model with broad cross-modal knowledge, followed by visual instruction tuning that enables the model to tackle diverse vision-centric tasks by framing them as language instructions. Recently, continual visual instruction tuning (CVIT) \cite{chen2024coin, wang2025smolora} has been proposed to sustain strong capability on downstream tasks by incrementally adapting the model to new task environments through sequential visual instruction tuning.

Despite the performance gains brought by CVIT, its implications for model safety have received limited attention. Current MLLMs remain vulnerable to harmful vision and language inputs \cite{ye2025survey}, raising critical safety concerns. To mitigate these risks, safety alignment \cite{zong2024safety, yin2025towards} has been introduced as an additional stage after pre-training and instruction tuning. Through techniques such as supervised fine-tuning \cite{zong2024safety, ding2025rethinking} or preference optimization \cite{zhang2025spa, weng2025adversary}, safety alignment ensures that MLLMs produce safe and reliable outputs even when exposed to unsafe visual or textual content. Nevertheless, existing works on CVIT \cite{he2023continual, zhu2024model} primarily focus on MLLMs before the safety alignment stage (pre-SA CVIT), while the effect of continual adaptation after safety alignment (post-SA CVIT) remains unexplored. In real-world deployment, continual updating after safety alignment is inevitable, making the concern for safety during CVIT both practical and urgent.

\begin{figure}[t]
  \centering
   \includegraphics[width=\linewidth]{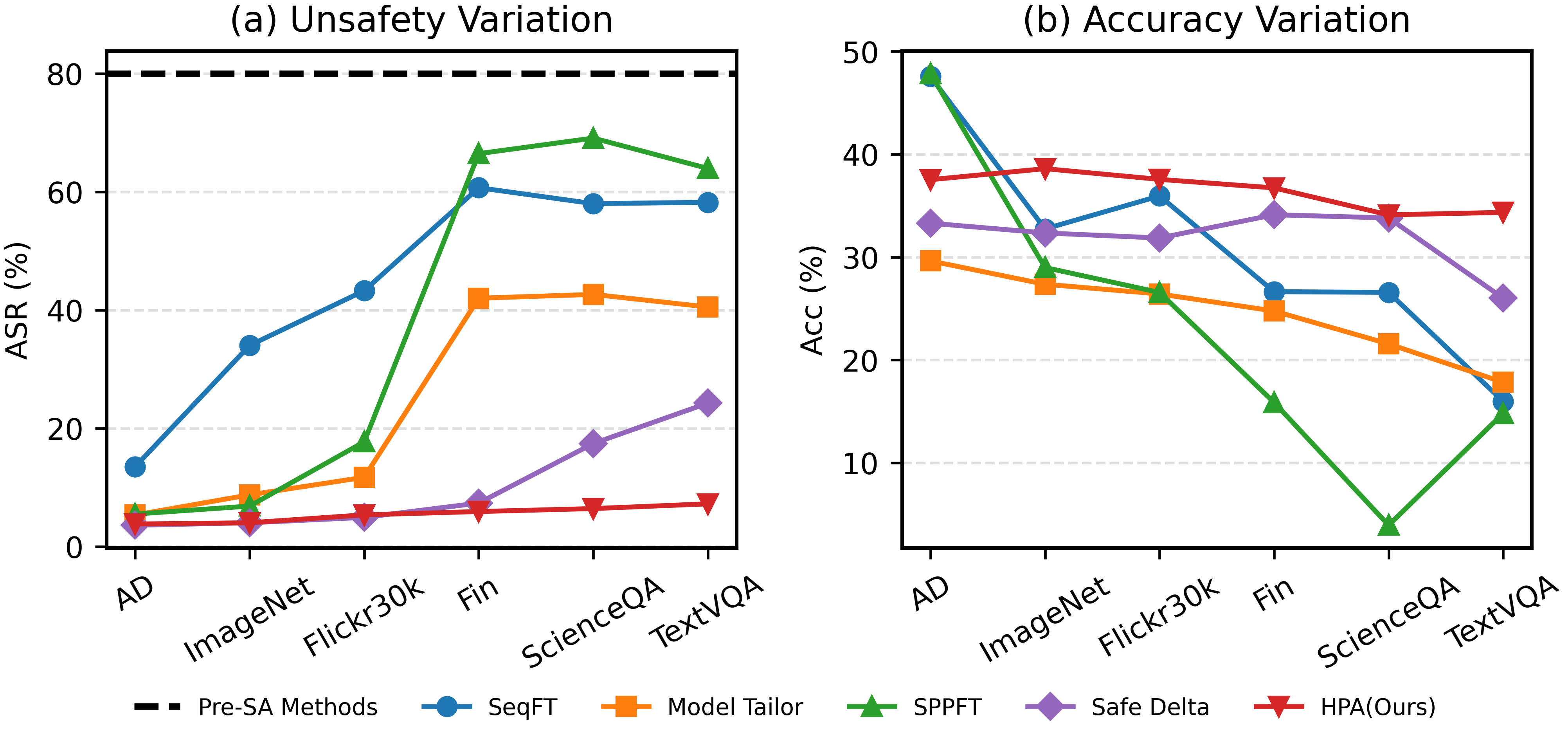}
   \caption{Unsafety Variation (on the safety benchmark) and Accuracy Variation (on the first tuning dataset AD) under Post-SA CVIT for different methods. The black dashed line indicates that the MLLM remains unsafe throughout Pre-SA CVIT due to missing safety alignment.}
   \label{decline}
   \vspace{-5mm}
\end{figure}

In this work, we pioneer the investigation of post-SA CVIT, examining how continual adaptation impacts safety-aligned MLLMs. Pre-SA CVIT is inherently limited to addressing task forgetting, entirely neglecting the dimension of safety. Suffering from the absence of safety alignment, the MLLM perpetuates its vulnerability across the continual adaptation steps in the pre-SA CVIT setting, as illustrated by the black dashed line in Figure~\ref{decline} (a). 
In contrast, the post-SA CVIT setting introduces a dual and more formidable challenge: as shown in Figure~\ref{decline}, MLLMs not only experience catastrophic forgetting but also exhibit a continual degradation of pre-established safety. 

Due to the coexistence of these challenges, developing an effective solution for post-SA CVIT is essential. Existing CVIT methods \cite{he2023continual, zhu2024model} and safety alignment approaches \cite{li2025safety, lusafe} struggle to achieve a harmonious balance between safety and capability during continual fine-tuning.
To address this problem, we propose \underline{H}armonious \underline{P}arameter \underline{A}daptation (HPA) in CVIT for MLLMs, an efficient post-training framework designed for the adaptation of safety alignment and capability learning.

HPA consists of focusing-based parameter partition, harmoniously balanced selection, and orthogonal parameter adjustment. To maintain the efficiency of MLLMs during CVIT, HPA updates model parameters only after training, without interfering with the existing training pipeline. At each fine-tuning step, HPA first partitions parameters into safety-focused and task-focused types based on their sensitivity differences before and after tuning, which enables a more effective selection of parameters that are crucial for either safety or capability. Subsequently, HPA achieves balanced parameter selection from both intra-layer and inter-layer perspectives, preventing one-sided parameter retention from disrupting the safety–capability trade-off. 
Given the above components lack a dedicated mechanism to alleviate catastrophic forgetting,
HPA further employs orthogonal parameter adjustment to ensure that adaptations for the current task minimally interfere with representations learned from previous tasks. Through carefully balanced parameter adaptation at each stage, HPA harmonizes the trade-off between capability learning and safety guarantee. Our contributions are summarized as follows:

\begin{itemize}[leftmargin=*]

\item We shift focus to post-SA CVIT, identifying that safety-aligned MLLMs suffer from task forgetting and safety degradation during CVIT.
\item We propose HPA for post-SA CVIT, a new post-training framework that selectively updates parameters via focusing-based partition, balanced selection, and orthogonal adjustment to achieve a harmonious balance between safety and task performance.
\item Extensive experiments on CVIT and safety benchmarks establish our method's superiority, achieving a better trade-off between robust safety assurance and preserved task performance than the compared baselines.
\end{itemize}

%% file: sec/2_related.tex
\section{Related Work}
\label{sec:related}
\textbf{CVIT for MLLMs.}
CVIT enhances MLLMs by sequentially fine-tuning them on a series of vision instruction datasets, enabling adaptation to evolving task environments. A number of benchmarks \cite{chen2024coin, wang2025smolora, zhao2025mllm} and methods \cite{he2023continual, zheng2024beyond, zhang2025enhancing, che2025lora} have been introduced to study this field. Early work such as Eproj \cite{he2023continual} first combined visual instruction tuning with continual learning and proposed a structure-expansion strategy to mitigate catastrophic forgetting. Fwd-Prompt \cite{zheng2024beyond} addressed negative forward transfer through a prompt-based intervention. CoIN \cite{chen2024coin} further established a comprehensive CVIT benchmark to evaluate MLLMs under sequential instruction tuning.

More recent approaches typically rely on additional parameter modules and modifications to the original training process. MoeLoRA \cite{chen2024coin} integrates a token-wise mixture-of-experts mechanism to selectively activate LoRA experts. SMoLoRA \cite{wang2025smolora} applies separable routing to alleviate dual forgetting. BranchLoRA \cite{zhang2025enhancing} introduces an asymmetric framework to capture both task-invariant and task-specific factors. SEFE \cite{chen2025sefe} tackles superficial forgetting via answer-style diversification and essential forgetting through constraining masked parameters to maintain key knowledge across tasks. LiLoRA \cite{che2025lora} expands lightweight task-specific structures for each new task. However, these methods increase parameter redundancy and training overhead, and they do not account for the safety alignment requirements of MLLMs. In contrast, our work focuses on addressing both task forgetting and safety degradation during CVIT without altering the original training pipeline, aiming to maintain efficiency while preserving safety.

\noindent \textbf{Safety Alignment for MLLMs.}
Safety alignment \cite{ji2023beavertails, zhou2024alignment} aims to ensure that model behavior remains consistent with human values, ethical norms, and intended usage constraints. With the rapid development of MLLMs, efforts toward safety alignment in multimodal settings have also grown. VLGuard \cite{zong2024safety} introduces a vision-language safety instruction-following dataset to align MLLMs toward safer responses. SPA-VL \cite{zhang2025spa} automatically constructs safety preference data to enable multimodal preference alignment. ADPO \cite{weng2025adversary} incorporates adversarial training into preference optimization to strengthen safety under worst-case adversarial perturbations. SEA \cite{lu2025sea} synthesizes learnable embeddings for additional modalities, making it possible to conduct multimodal safety alignment even when only textual safety data is available. However, repeatedly performing safety alignment during CVIT is constrained by privacy concerns and prohibitive computational costs. In this work, we explore a more efficient post-training strategy to maintain safety while preserving downstream task performance during continual adaptation.

%% file: sec/3_method.tex
\section{Method}
\label{sec:method}

\subsection{Problem Formulation}

Given a MLLM $f(x;\theta)$, CVIT seeks to incrementally fine-tune $f$ to acquire a sequence of vision tasks 
$\mathcal{T} = \{\tau_1, \tau_2, \ldots, \tau_n\}.$
At each time step $t$, the model is exposed to a new task $\tau_t$ together with its associated dataset $\mathcal{D}_t$, in order to integrate the new knowledge into its existing capabilities. Typically, the dataset $\mathcal{D}_t$ follows the format $\{X^{ins}, X^{vis}, X^{ans}\}$, where $X^{ins}$, $X^{vis}$, and $X^{ans}$ denote the textual instruction input, the visual input, and the linguistic answer, respectively. For pre-SA CVIT, the final optimization objective can be formulated as:
\begin{equation}
\min_{\{\theta_t\}_{t=1}^n}\;
\sum_{t=1}^{n}\Bigg(
    \sum_{i=1}^{t}\mathcal{L}_{C_i}\!\big(f(x;\theta_t)\big)
\Bigg),
\end{equation}
where $\mathcal{L}_{C_i}$ denotes the performance loss for the $i$-th task $\tau_i$.  This objective focuses solely on the performance changes of the current task $\tau_t$ and the previous $t-1$ tasks.

Now consider a safety-aligned MLLM $f(x;\theta_0)$. After the first task-specific tuning, the model becomes $f(x;\theta_1)$. Due to privacy concerns and high resource consumption, it is infeasible to repeatedly re-align the model with the original safety dataset during CVIT. Under this constraint, the key challenge for post-SA CVIT is to prioritize preserving safety while minimizing the impact on both current-task learning and previously learned tasks.

Formally, during post-SA CVIT at each step $t$, the optimization objective can be expressed as:
\begin{equation}
\min_{\{\theta_t\}_{t=1}^n}\;
\sum_{t=1}^{n}\Bigg(
    \mathcal{L}_S\!\big(f(x;\theta_t)\big)
    \;+\;
    \sum_{i=1}^{t}\mathcal{L}_{C_i}\!\big(f(x;\theta_t)\big)
\Bigg),
\end{equation}
where $\mathcal{L}_S$ denotes the safety loss.

\subsection{Overview of HPA Framework}

Assume that at time step $t$, we have the fine-tuned model $f(x; \theta_t)$ and the pre-tuned model $f(x; \theta_{t-1})$. Our goal is to design a post-training parameter adaptation scheme that achieves an optimal balance between safety and task performance. Specifically, the final adapted parameter at the $l$-th layer is defined as
\begin{equation}
\hat{W}_t^l = \mathcal{F}(W_{t-1}^l, W_t^l),
\end{equation}
where $W_{t-1}^l, W_t^l \in \mathbb{R}^{r\times c}$ denote the weights of the $l$-th layer in $f(x; \theta_{t-1})$ and $f(x; \theta_t)$, respectively.

\begin{figure*}[t]
  \centering
   \includegraphics[width=\linewidth]{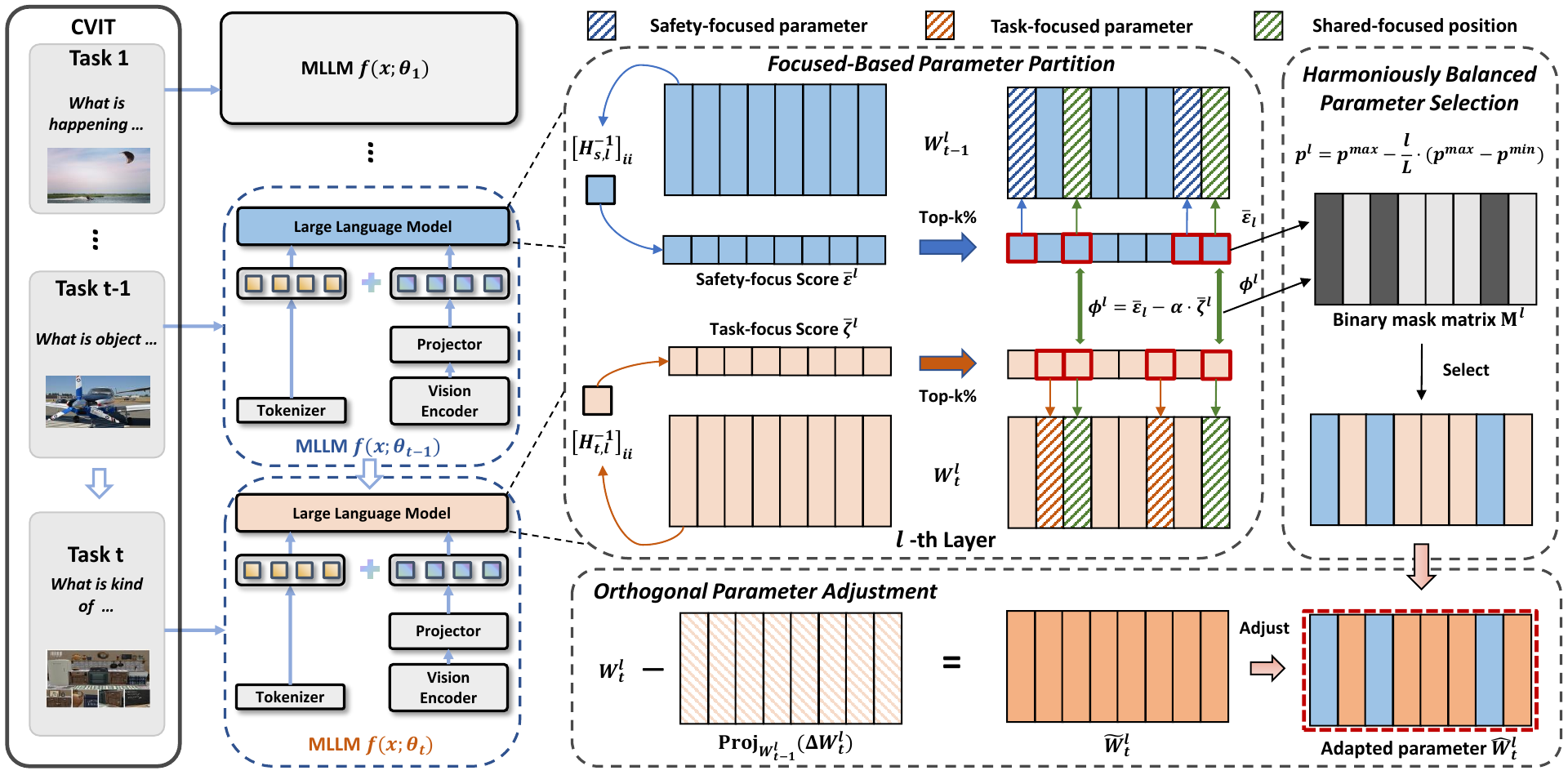}
   \caption{Overview of HPA Framework. HPA consists of three components: (1) \textbf{Focusing-based parameter partition} which partitions parameters into safety- and task-focused types; (2) \textbf{Harmoniously balanced parameter selection} which balances the selection of focused parameters; (3) \textbf{Orthogonal parameter adjustment} which applies orthogonal constraints during parameter updates to mitigate forgetting.}
   \label{fig_main}
\end{figure*}

To achieve this goal, our framework is built upon three key components:
(1) \textbf{Focusing-based parameter partition}, which partitions the parameters in $W_{t-1}^l$ and $W_t^l$ based on their distinct focuses on safety preservation and task adaptation (Sec~\ref{sec3.3});
(2) \textbf{Harmoniously balanced parameter selection}, which harmonizes old and new parameters via layer-wise safety-focused selection and dynamically adjusted retention across layers (Sec~\ref{sec3.4}); and
(3) \textbf{Orthogonal parameter adjustment}, which constrains parameter updates through orthogonal optimization to further mitigate the problem of catastrophic forgetting (Sec~\ref{sec3.5}). The overall framework of HPA is shown in Figure~\ref{fig_main}.

\subsection{Focusing-Based Parameter Partition}
\label{sec3.3}

Due to the inherent over-parameterization of deep neural networks, not all parameters contribute equally to fitting the target distribution \cite{huanglearn}. In other words, certain parameters are more essential in knowledge representation or safety preservation, while others remain relatively redundant.
In this work, parameters are considered focused when they strongly contribute to either safety or task performance. Accordingly, we partition two types of focused parameters:
(1) \textbf{Safety-focused parameters}, defined as the top-$k\%$ parameters in $W_{t-1}^l$ that contribute most to safety preservation; and 
(2) \textbf{Task-focused parameters}, defined as the top-$k\%$ parameters in $W_t^l$ that are most important for the current task.

Each parameter’s focus is primarily determined by its relative importance in a particular scenario. Previous methods \cite{han2015learning, huanglearn} that estimate parameter importance through magnitude or gradient variations are often too coarse-grained to accurately capture this focusing behavior.
Inspired by Hessian-based pruning methods \cite{frantar2022spdy, frantar2023sparsegpt}, we estimate the importance of each parameter $w_{i,j}$ by the loss increase incurred upon its removal, given by
${(w_{i,j})^2}/{\big[\mathrm{H}^{-1}\big]_{jj}}$, where $\mathrm{H}$ denotes the Hessian of the loss with respect to the model parameters,
$\mathrm{H}^{-1}$ is its inverse, and $\big[\mathrm{H}^{-1}\big]_{jj}$ represents the $j$-th diagonal element of $\mathrm{H}^{-1}$.

This metric captures the sensitivity of model performance to the removal of each parameter. During fine-tuning, we can analogously evaluate a parameter’s importance based on its variation before and after tuning. Specifically, for each entry $(i,j)$ of $W_{t-1}^l$ and $W_t^l$, we define two importance scores in layer $l$: the safety-focus score $\varepsilon$ and task-focus score $\zeta$:
\begin{equation}
\label{Hessian}
\begin{aligned}
\varepsilon_{i,j}^l &= \frac{\big(W_{t-1}^l(i,j) - W_t^l(i,j)\big)^2}{\big[\mathrm{H}_{s,l}^{-1}\big]_{ii}}, \\
\zeta_{i,j}^l &= \frac{\big(W_{t}^l(i,j) - W_{t-1}^l(i,j)\big)^2}{\big[\mathrm{H}_{t,l}^{-1}\big]_{ii}},
\end{aligned}
\end{equation} 
where $\mathrm{H} = 2X^{\top}X$, and $X \in \mathbb{R}^{d\times r}$ is the activation matrix computed from the corresponding safety or task calibration dataset $\mathcal{D}^*_s$ and $\mathcal{D}^*_t$.

After obtaining $\varepsilon_{i,j}^l$ and  $\zeta_{i,j}^l$, we aggregate the importance scores column-wise by averaging across $r$ rows: 
\begin{equation}
\label{Hessian2}
\bar{\varepsilon}^l_j = \frac{1}{{r}} \sum_{i=1}^{{r}} \varepsilon_{i,j}^l, \quad  
\bar{\zeta}^l_j = \frac{1}{{r}} \sum_{i=1}^{{r}} \zeta_{i,j}^l, 
\end{equation}
where the resulting vectors $\bar{\varepsilon}^l, \bar{\zeta}^l \in \mathbb{R}^{c}$ indicate the column-wise focusing strength in layer $l$. By selecting the top-$k\%$ indices from $\bar{\varepsilon}^l$, we can identify the safety-focused and task-focused parameters in the parameter space.

However, a fundamental challenge arises from the overlap phenomenon between the two focused parameter sets. Specifically, for a given position $(i,j)$ in $\hat{W}_t^l$, if the corresponding column in $W_{t-1}^l$ belongs to the safety-focused set while the same position in $W_t^l$ belongs to the task-specific set, these positions are defined as the \textbf{shared-focused positions}, which jointly influence both safety and task performance.
Therefore, in addition to preserving safety-focused parameters, the selection process must also decide whether to retain $W_{t-1}^l$ or $W_t^l$ parameters at the shared-focused positions.

\subsection{Harmoniously Balanced Parameter Selection}
\label{sec3.4}
During the CVIT process, after each task-specific fine-tuning step, we aim for the updated model $f(x; \theta_t)$ to retain those parameters from the previous model $f(x; \theta_{t-1})$ that are most critical for maintaining safety, while minimally affecting parameters essential to the current task. Thus, the primary goal of parameter adaptation is to determine how to select parameters from $W_{t-1}^l$ and $W_t^l$ to achieve an optimal balance between model safety and performance on the current task. Let $\mathrm{M}^l \in \mathbb{R}^{r \times c}$ be a binary mask matrix, the parameter selection process can be formulated as:
\begin{equation}
    \hat{W}_t^l = \mathrm{M}^l \odot W_{t-1}^l + (\mathrm{1} - \mathrm{M}^l) \odot W_t^l,
\end{equation}
where $p\%$ of the columns in $\mathrm{M}^l$ are set to $1$, indicating the parameters retained from $W_{t-1}^l$. In practice, $p$ is typically set below the focused-parameter ratio $k$, ensuring selected parameters form a tighter subset of the initial focused set.

For the $l$-th layer, the most straightforward strategy is to directly retain $p\%$ of the safety-focused parameters. However, such a naive selection may discard task-focused parameters located in shared-focused positions, thereby interfering with the current task performance. To avoid this interference, we first retain from $W_{t-1}^l$ the safety-focused parameters that do not lie in the shared-focused positions, accounting for $p_s\%$ of all parameters. Considering that the number of safety-focused parameters may not reach the desired proportion $p$, we further select a portion of parameters located at the shared-focused positions from $W_{t-1}^l$ to be preserved.

To determine the relative contribution of these parameters to safety and task performance, we define a balancing score $\phi^l \in \mathbb{R}^{\lfloor c \cdot (k\%-p_s\%) \rfloor}$ for shared-focused positions as
\begin{equation}
\label{phi}
\phi^{l} = \bar{\varepsilon}^{l} - \alpha \cdot \bar{\zeta}^{l},
\end{equation}
where $\alpha$ adaptively balances the contributions of safety and task importance during shared-focused parameter selection, is defined as:
\begin{equation}
\label{eq:alpha}
\alpha = \alpha_{1} - \tfrac{1}{2}(\alpha_{1} - \alpha_{0}) \left[\tanh\!\big(\mathbb{E}\!\left[\log\!\tfrac{\bar{\varepsilon}^l}{\bar{\zeta}^l}\right]\big) + 1\right],
\end{equation}
where $\alpha_{0}$ and $\alpha_{1}$ denote the lower and upper bounds of $\alpha$, respectively; $\mathbb{E}[\cdot]$ represents the expectation operator; and $\tanh(\cdot)$ denotes the hyperbolic tangent function.
A larger $\phi^{l}$ indicates a stronger safety-focus tendency. Based on $\phi^{l}$, we select the top-$(p - p_s)\%$ indices from all $c$ columns as the remaining safety parameters from $W_{t-1}^l$.

During parameter selection, the retention ratio $p$ is also determined to control how many parameters of $W_{t-1}^l$ are preserved. Increasing $p$ improves safety retention but may interfere more with current task adaptation. Given the inherent heterogeneity across layers, our method adaptively adjusts $p$ according to the layer index. During fine-tuning of large models, higher layers closer to the output typically encode more task-specific knowledge \cite{ju2024large}; hence we assign a smaller retention ratio to deeper layers to favor current-task adaptation over strict preservation of prior safety constraints. Formally, $p^l$ decreases linearly with layer depth:
\begin{equation}
\label{p^l}
p^l = p^{max} - \frac{l}{L} \cdot (p^{max} - p^{min}),
\end{equation}
where \( p^{max} \) and \( p^{min} \) denote the parameter retention ratios for the first and last layers, respectively, and \( L \) is the total number of layers.

\subsection{Orthogonal Parameter Adjustment}
\label{sec3.5}
The harmoniously balanced parameter selection effectively balances model safety and task performance at step $t$ by selectively retaining focused parameters from $W_{t-1}^l$. However, for the parameters retained from $W_t^l$, the update process should avoid interfering with previously learned representations. Therefore, we enforce an approximate orthogonality between the new update directions and the subspace spanned by past parameters, effectively reducing catastrophic forgetting during continual adaptation. At step $t$, the parameter update in layer $l$ is defined as
\begin{equation}
\Delta W_t^l = W_t^l - W_{t-1}^l.
\end{equation}

To reduce interference with past knowledge, we perform an orthogonal parameter update, ensuring that the new update direction remains as independent as possible from the existing parameter space. Specifically, we first compute the projection of the update $\Delta W_t^l$ onto the previous parameters $W_{t-1}^l$:
\begin{equation}
\text{Proj}_{W_{t-1}^l}(\Delta W_t^l) = 
\frac{\langle \Delta W_t^l, W_{t-1}^l \rangle}{\|W_{t-1}^l\|_F^2} \, W_{t-1}^l,
\end{equation}
where $\langle \cdot, \cdot \rangle$ denotes the Frobenius inner product and $\|\cdot\|_F$ the Frobenius norm. $\text{Proj}_{W_{t-1}^l}(\Delta W_t^l)$ represents the projection of the current update $\Delta W_t^l$ onto the subspace spanned by the previous parameters $W_{t-1}^l$. This term quantifies the component of the new update that aligns with previously learned knowledge.
By subtracting this projection, we remove the part of the update that interferes with old task representations, ensuring that the resulting update is orthogonal to $W_{t-1}^l$:
\begin{equation}
\label{update}
\tilde{W_t^l}= W_{t-1}^l + \Delta W_t^l - \text{Proj}_{W_{t-1}^l}(\Delta W_t^l).
\end{equation}
The final parameter adaptation can be formulated as:
\begin{equation}
\label{final}
 \hat{W}_t^l = \mathrm{M}^l \odot W_{t-1}^l + (\mathrm{1} - \mathrm{M}^l) \odot \tilde{W_t^l}.
\end{equation}

The overall implementation procedure for HPA under CVIT is summarized in Algorithm~\ref{alg:scd}.

\begin{algorithm}[tb]
\caption{Implementation of HPA}
\label{alg:scd}
\textbf{Input}: Tuning datasets $\mathcal{D}=\{\mathcal{D}_1,\mathcal{D}_2,\dots,\mathcal{D}_n\}$; safety-aligned MLLM $f(x;\theta_0)$; calibration set $\mathcal{D}^*=\{\mathcal{D}_s^*,\mathcal{D}_1^*,\mathcal{D}_2^*,\dots,\mathcal{D}_n^*\}$ \\
\textbf{Output}: Adapted MLLM $f(x;\theta_n)$

\begin{algorithmic}[1]
\FOR{$t=1$ \TO $n$}
    \STATE \textbf{Train} $f(x;\theta_t)$ on $\mathcal{D}_t$ initialized from $f(x;\theta_{t-1})$
    \FOR{each layer $l$ of $f$}
        \STATE Extract weights $W_{t-1}^l$ and $W_t^l$
        \STATE Compute importance scores $\bar{\varepsilon}^l$, $\bar{\zeta}^l$ using Eq.~\ref{Hessian} and Eq.~\ref{Hessian2} with $\mathcal{D}_s^*$ and $\mathcal{D}_t^*$
        \STATE Derive $\phi^l$ and $p^l$ via Eq.~\ref{phi} and Eq.~\ref{p^l}
        \STATE Select parameters using $\bar{\varepsilon}^l$, $\phi^l$, $p^l$ to form the binary mask matrix $\mathrm{M}^l$
        \STATE Compute orthogonal update $\tilde{W}_t^l$ using Eq.~\ref{update}
        \STATE Obtain adapted weights $\hat{W}_t^l$ via Eq.~\ref{final}
        \STATE Update $\theta_t$ by replacing $W_t^l$ with $\hat{W}_t^l$
    \ENDFOR
\ENDFOR
\STATE \textbf{return} $f(x;\theta_n)$
\end{algorithmic}
\end{algorithm}

%% file: sec/4_exp.tex
\section{Experiments}
\label{sec:exp}

\subsection{Benchmark Setup}
In this section, we outline the benchmark setup used in our study. To evaluate how safety-aligned MLLMs evolve during the CVIT process, we construct a continual training and testing pipeline with six visual instruction tuning datasets and measure safety performance using two established benchmarks.

\textbf{CVIT Datasets.} We employ a diverse set of datasets covering VQA, image classification, and visual reasoning tasks: AD \citep{sima2024drivelm}, ImageNet \citep{deng2009imagenet}, Flickr30k \citep{plummer2015flickr30k}, Fin \citep{zhao2025mllm}, ScienceQA \citep{lu2022learn}, and TextVQA \citep{singh2019towards}. This variety ensures broad task coverage in continual adaptation.

\textbf{Safety Datasets.} Safety is assessed with VLGuard \citep{zong2024safety} and Ch3EF \citep{shi2024assessment}, two widely used benchmarks for multimodal alignment. For VLGuard, we focus on two subsets: VLGuard-safe-unsafe (VLG-1) and VLGuard-unsafe (VLG-2) to capture distinct types of safety risks.

\textbf{Metrics.}
Let $a_{k,j}$ denote the accuracy on the $j$-th task after fine-tuning on the $k$-th task. We evaluate model performance using \textbf{Average Performance (AP)} and \textbf{Backward Transfer (BWT)} \citep{wang2024comprehensive}, which respectively measure the overall accuracy after training on task $k$ and the degree of forgetting on previous tasks:  

\begin{equation}
\mathrm{AP}_{k}=\frac{1}{k} \sum_{j=1}^{k} {a}_{k, j}, \;
\mathrm{BWT}_{k}=\frac{1}{k-1} \sum_{j=1}^{k-1} ({a}_{k, j}-{a}_{j,j}).
\label{formula_AP}
\end{equation}

For safety evaluation, \textbf{Attack Success Rate (ASR)} \citep{shi2024assessment} is defined as the proportion of harmful inputs for which the model fails to provide a safe response. A lower ASR indicates stronger safety. Based on this metric, we introduce two measures: \textbf{MASR} and \textbf{DASR}. After fine-tuning on the $k$-th task, MASR represents the average ASR across three safety datasets, while DASR quantifies the increase in MASR relative to the initial safety-aligned model.

\begin{table*}[htb!]
\caption{Performance metrics (\%) on the CVIT and safety benchmarks, evaluated after completing the final tuning task. The table displays per-task accuracy on the six CVIT datasets, alongside task performance (AP, BWT) and safety performance (MASR, DASR).}
\label{table:main-results}

\centering
\resizebox{\textwidth}{!}{
\renewcommand{\arraystretch}{1.5}
\begin{tabular}{l || c c c c c c | c c || c c c | c c}
\toprule[1.2pt]
\multirow{4}{*}{\textbf{Method}} & \multicolumn{8}{c||}{\textbf{CVIT Task Performance ($\uparrow$)}} & \multicolumn{5}{c}{\textbf{Safety Performance ($\downarrow$)}} \\

\cmidrule(lr){2-9}\cmidrule(lr){10-14}

& \multicolumn{6}{c|}{\textbf{Dataset}} & \multicolumn{2}{c||}{\textbf{Metric}} & \multicolumn{3}{c|}{\textbf{Dataset}} & \multicolumn{2}{c}{\textbf{Metric}} \\

\cmidrule(lr){2-7}\cmidrule(lr){8-9}\cmidrule(lr){10-12}\cmidrule(lr){13-14}

& {\textbf{AD}} & {\textbf{ImageNet}} & {\textbf{Flickr30k}} & {\textbf{Fin}} & {\textbf{ScienceQA}} & {\textbf{TextVQA}} & {\textbf{AP}} & {\textbf{BWT}} 
& {\textbf{VLG-1}} & {\textbf{VLG-2}} & \textbf{Ch3EF} & {\textbf{MASR}} & {\textbf{DASR}} \\

\midrule 

Zero-shot & 18.95 & 20.63 & 6.55 & 0.04 & 24.50 & 0.02 & 11.78 & - 
& 0.18 & 2.04 & 6.37 & 2.86 & -  \\
DirFT & 49.63 & 93.96 & 152.60 & 86.56 & 83.42 & 56.60 & 87.13 & - 
& - & - & - & - & -  \\
\cmidrule{1-14} 

\multicolumn{14}{c}{\textbf{{Original Data}}} \\ \cmidrule{1-14}
SeqFT & 19.32 & 45.19 & 106.64 & 85.23 & 81.13 & 56.60 & 65.68 & -25.62 
& 76.88 & 25.34 & 25.46 & 42.56 & 39.70  \\
EWC & 19.52 & 46.08 & 106.51 & 84.77 &\textbf{81.70}& 56.53 & 65.85 & -24.90 
& 76.70 & 24.89 & 25.46 & 42.35 & 39.49  \\
{Replay} & 21.33 & 25.64 & 141.96 & \textbf{85.32} & 81.41 & \textbf{59.85} & 69.25 & -21.40 
& 60.39 & 12.90 & 19.92 & 31.07 & 28.21 \\
SEFE & 17.92 & 41.01 & 127.27 & 79.71 & 78.89 & 54.19 & 66.49 & -20.83 
& 79.75 & 22.17 & 25.46 & 42.46 & 39.60 \\ 
Model Tailor & 20.32 & 35.88 & \textbf{145.76} & 79.04 & 78.89 & 52.84 & 68.79 & -10.29 
& 60.22 & 7.92 & 16.73 & 28.29 & 25.43 \\
SPPFT & 16.26 & 43.86 & 110.34 & 78.22 & 63.97 & 53.7 & 61.06 & -29.81 
& 77.96 & 47.96 & 13.14 & 46.35 & 43.49 \\
Safe Delta & 33.67 & 46.36 & 142.10 & 83.78 & 80.35 & 53.67 & 73.32 & -6.91 
& 1.25 & 4.56 & 9.24 & 5.02 & 2.15 \\
\cmidrule{1-14}
 \rowcolor[gray]{0.9} \textbf{HPA (Ours)} & \textbf{35.01} & \textbf{60.48} & 144.80 & {82.73} & 78.99 & {52.36} & \textbf{75.73} & \textbf{-4.87} 
& \textbf{0.72} & \textbf{4.30} & \textbf{9.24} & \textbf{4.75} & \textbf{1.89} \\
\cmidrule{1-14} 

\multicolumn{14}{c}{\textbf{0.1\% Harmful Data Injected}} \\ \cmidrule{1-14}
SeqFT & 15.98 & 50.24 & 111.39 & 84.07 & 81.58 & \textbf{56.86} & 66.69 & -24.29
& 81.54 & 45.70 & 47.43 & 58.22 & 55.36  \\
EWC & 16.04 & 43.41 & 109.98 & 83.23 & 80.92 & 56.06 & 64.94 & -25.81 
& 83.15 & 47.96 & 47.23 & 59.45 & 56.58  \\
{Replay} & 23.48 & 59.27 & 129.41 & \textbf{86.46} & \textbf{82.05} & 56.74 & 72.90 & -17.89 
& 77.24 & 42.31 & 46.82 & 55.46 & 52.59 \\
SEFE & 14.81 & 41.37 & 126.72 & 81.18 & 78.89 & 54.17 & 66.19 & -20.59 
& 82.62 & 27.60 & 30.18 & 46.80 & 43.94 \\ 
Model Tailor & 17.83 & 42.04 & \textbf{155.64} & 77.06 & 78.83 & 54.03 & 70.91 & -11.79 
& 67.20 & 15.16 & 39.19 & 40.52 & 37.65 \\
SPPFT & 14.80 & 43.82 & 105.76 & 68.96 & 68.85 & 53.80 & 59.33 & -31.21 
& 93.91 & 80.09 & 17.86 & 63.95 & 61.09 \\
Safe Delta & 26.01 & 47.98 & 146.82 & 83.80 & 81.04 & 53.54 & 73.20 & -6.82 
& 50.90 & 5.66 & 16.22 & 24.26 & 21.40 \\
\cmidrule{1-14}
\rowcolor[gray]{0.9} \textbf{HPA (Ours)} & \textbf{34.35} & \textbf{63.49} & 148.62 & {82.13} & 78.76 & 52.34 & \textbf{76.62} & \textbf{-3.88} 
& \textbf{1.43} & \textbf{5.66} & \textbf{14.58} & \textbf{7.22} & \textbf{4.36} \\
\bottomrule[1.2pt]
\end{tabular}
}
\end{table*}

\subsection{Experimental Setup}
\textbf{Comparison Methods.} We compare our approach against a broad range of baselines to highlight its superior performance. Zero-shot denotes the performance of the safety-aligned model without any task-specific fine-tuning, while DirFT refers to directly fine-tuning the model on the corresponding visual instruction dataset. SeqFT represents sequential fine-tuning across CVIT datasets. The remaining methods fall into three categories: (1) classical continual learning methods like EWC \citep{kirkpatrick2017overcoming} and Replay \citep{chaudhry2019tiny}; (2) recent CVIT approaches, including Model Tailor \citep{zhu2024model} and SEFE \citep{chen2025sefe}; and (3) safety-protected techniques for LLMs, such as SPPFT \citep{li2025safety} and Safe Delta \citep{lusafe}.

\noindent \textbf{Calibration Sets Construction.} Post-training methods typically require a small calibration set for parameter refinement. We denote the calibration set as $\mathcal{D}^* = \{{X^{ins}, X^{vis}, X^{ans}}\}$, where each triplet consists of an instruction, a visual input, and a corresponding answer. For each downstream task $t$, the task-specific calibration set $\mathcal{D}^*_t$ is constructed by randomly sampling a small number of instances from the corresponding fine-tuning dataset.

For safety alignment, direct access to the original alignment data is often restricted due to privacy concerns, and only the aligned model $f(x; \theta_0)$ is available. To circumvent this limitation, we construct a safety calibration set by exploiting the inherent safety behavior of $f(x; \theta_0)$. Following the four major safety risk categories \citep{zong2024safety}: Privacy, Risky Behavior, Deception, and Hateful Speech, we collect a small set of harmful visual samples $X^{vis}_{unsafe}$ and pair them with corresponding unsafe instructions $X^{ins}_{unsafe}$. Since $f(x;\theta_0)$ is safety-aligned, it produces safe responses $X^{ans}_{safe}$, forming  
$\mathcal{D}^*_s = \{X^{ins}_{unsafe}, X^{vis}_{unsafe}, X^{ans}_{safe}\}.$

\noindent \textbf{Implementation Details.} We use the pretrained first-stage LLaVA-v1.5-7B \cite{liu2024visual} as our base model to avoid potential data leakage issues in CVIT. During the safety alignment phase, we align the model using the VLGuard and SPA-VL \cite{zhang2025spa} datasets. In the CVIT phase, LoRA \cite{hu2022lora} is chosen as the fine-tuning method and is merged into the base model after fine-tuning. The hyperparameter settings for our method are as follows: the number of samples in the calibration sets $\mathcal{D}^*_s$ and $\mathcal{D}^*_t$ are 8 and 128, respectively; the values of $\alpha_0$, $\alpha_1$, $p_{min}$, and $p_{max}$ are set to 0.4, 0.8, 5, and 15, respectively; and $k=2p^l$. Our method is applied to all linear layers of the base MLLM.

\subsection{Main Results}
We assess HPA against baseline methods by performing sequential fine-tuning on tasks from AD to TextVQA under two data conditions: the \textbf{Original Data} and the datasets with \textbf{0.1\% Harmful Data Injected}. Upon completing the final tuning task, we evaluate both the CVIT task performance and the safety performance of the models.

As shown in Table~\ref{table:main-results}, in the \textbf{Original Data} condition, HPA consistently outperforms all baseline methods after the final tuning task. Compared to Safe Delta, HPA not only improves CVIT task performance, with +2.41\% in AP and +2.04\% in BWT, but also enhances safety performance, with a reduction of 0.27\% in MASR and 0.26\% in DASR. These results demonstrate that HPA maintains high safety performance while effectively mitigating forgetting.
In the more challenging \textbf{0.1\% Harmful Data Injected} condition, HPA also achieves superior performance over all baselines. On CVIT task performance, HPA outperforms Safe Delta by +3.42\% in AP and +2.94\% in BWT. Notably, Safe Delta's safety performance degrades significantly under this condition, with attack success rates of 24.26\% MASR and 21.40\% DASR. In contrast, HPA achieves significantly better safety performance of 7.22\% MASR and 4.36\% DASR. These results highlight the robustness of HPA in more challenging scenarios.

\subsection{Ablation Study}
\noindent \textbf{Effect of Key Components in HPA. }
We conduct ablation studies on the three key components of HPA, as shown in Table~\ref{tab_ablation}. 
Specifically, we evaluate the impact of different components: retaining safety-focused parameters using the ${\varepsilon}_l$ score, 
preserving parameters at shared-focused positions based on ${\phi}_l$, 
and the orthogonal parameter adjustment resulting in $\tilde{W}_t^l$. Comparing Exp.2 with Exp.1, directly preserving the top $p\%$ of safety-focused parameters substantially enhances model safety. Exp.3 retains only the safety-focused parameters located at the shared-focused positions, leading to improved task performance but with a noticeable drop in safety. Exp.4 combines the strategies of Exp.2 and Exp.3, achieving better task performance while maintaining safety. Furthermore, Exp.5 introduces the orthogonal parameter adjustment constraint based on Exp.4, which effectively mitigates catastrophic forgetting during CVIT.

\begin{table}[t]
  \centering
  \caption{Ablation studies of three key components in HPA.}
    \resizebox{\columnwidth}{!}{
    \begin{tabular}{c|ccc|cccc}
    \toprule[1.0pt]
   
   Exp.  &{${\bar\varepsilon}^l$} &{${\phi}^l$} &{$\tilde{W_t^l}$}    
   &AP $\uparrow$ &BWT $\uparrow$ &MASR $\downarrow$ &DASR $\downarrow$ \\ \midrule[0.8pt]
  1 &$\times$ &$\times$ &$\times$  &66.69  &-24.29 &58.22 &55.36 \\
  2 &$\checkmark$ &$\times$ &$\times$  &73.49  &-5.81 &\textbf{6.02} &\textbf{3.16} \\
  3 &$\times$ &$\checkmark$ &$\times$ &74.16  &-5.21 &11.51 &8.64\\
  4 &$\checkmark$ &$\checkmark$ &$\times$ &74.82  &-7.00 &9.67 & 6.81\\
  \rowcolor[gray]{0.9} 5 &$\checkmark$ &$\checkmark$ &$\checkmark$  &\textbf{76.62}  &\textbf{-3.88} &\underline{7.22} &\underline{4.36}\\ \bottomrule[1.0pt]
    \end{tabular}}
  \label{tab_ablation}
   \vspace{-5mm}
\end{table}

\begin{figure}[t]
  \centering
   \includegraphics[width=\linewidth]{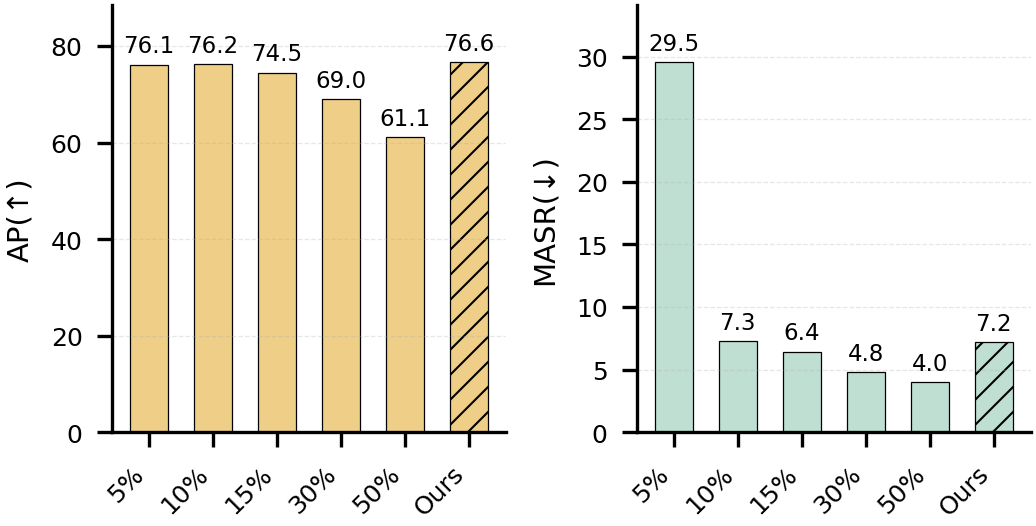}

   \vspace{-2mm}
   \caption{Effect of different parameter retention rates $p$ on performance and safety.
   }
   \label{fig_abp}
\end{figure}

\noindent \textbf{Effect of Retention Rate $p$. }
Figure~\ref{fig_abp} illustrates the impact of different parameter retention rates $p$ on model performance and safety during CVIT. 
The value of $p$ controls the proportion of safety-focused parameters being preserved. As $p$ increases, model safety consistently improves, while task performance gradually declines. When $p$ is relatively small (5\%--15\%), the performance degradation remains limited; however, from the safety perspective, allocating at least 10\% of safety-focused parameters is necessary to maintain robustness during continual adaptation. Moreover, our layer-wise dynamic retention ratio $p_l$ achieves a better trade-off by ensuring safety while enhancing task learning performance.

\begin{figure}[t]
  \centering
   \includegraphics[width=\linewidth]{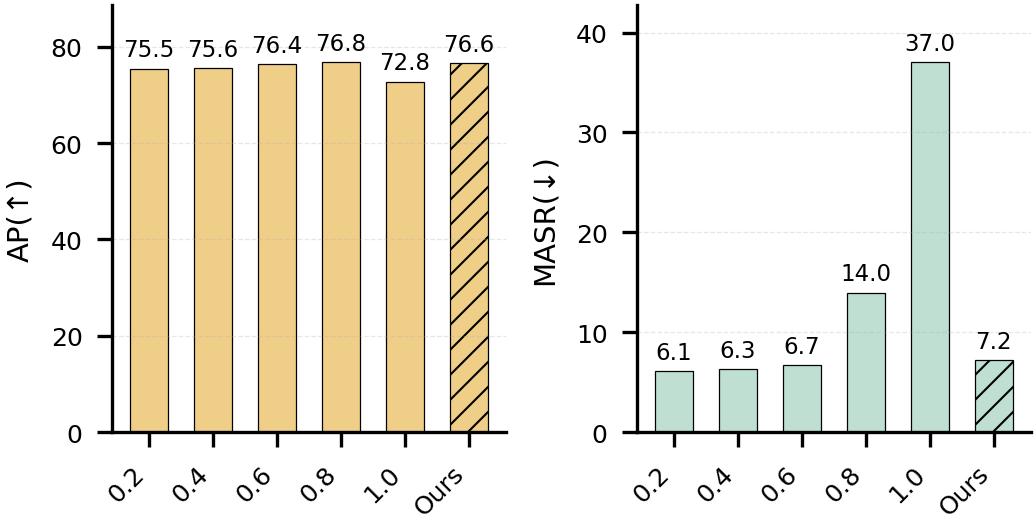}
   \vspace{-6mm}
   \caption{Effect of fixed versus adaptive coefficient $\alpha$ on performance and safety.
   }
   \label{fig_abbs}
   \vspace{-5mm}
\end{figure}

\noindent \textbf{Effect of Coefficient $\alpha$. }
We show the impact of the coefficient $\alpha$ on both model performance and safety. When $\alpha = 0$, the influence of the task-focused score on selection is effectively ignored. As shown in Figure~\ref{fig_abbs}, increasing a fixed $\alpha$ gradually shifts the model’s attention toward task performance, at the cost of a slight decrease in safety. However, when $\alpha$ becomes too large, the overall performance deteriorates significantly. Our method adaptively adjusts $\alpha$ for each layer based on the relative impact of safety and task signals, thereby maintaining a balanced trade-off.

\begin{table}[t]

  \centering
  \caption{Experimental results on different CVIT task orders.}
    \resizebox{\columnwidth}{!}{
    \begin{tabular}{c|c|cccc}
    \toprule[1.2pt]
   
   Order  &Method &AP $\uparrow$ &BWT $\uparrow$ &MASR $\downarrow$ &DASR $\downarrow$ \\ \midrule[0.8pt]
  \multirow{3}{*}{Order1} &SeqFT  &85.17  &-11.76 &62.49 &59.63 \\
  &Model Tailor  &78.00  &-3.23 &49.72 &46.86 \\
  & \cellcolor[gray]{0.9} Ours  & \cellcolor[gray]{0.9} \underline{79.45}  &\cellcolor[gray]{0.9} \textbf{3.64} &\cellcolor[gray]{0.9} \textbf{7.39} &\cellcolor[gray]{0.9} \textbf{4.52} \\ \cmidrule{1-6}
  \multirow{3}{*}{Order2} &SeqFT  &74.94  &-13.90 &61.90 &59.04 \\
  &Model Tailor  &71.00  &-8.82 &37.00 &34.14 \\
  &\cellcolor[gray]{0.9} Ours  &\cellcolor[gray]{0.9} \textbf{77.21}  &\cellcolor[gray]{0.9} \textbf{-4.72} &\cellcolor[gray]{0.9} \textbf{5.52} & \cellcolor[gray]{0.90} \textbf{2.66} \\ \bottomrule[1.2pt]
    \end{tabular}}
  \label{tab_order}
  
\end{table}

\noindent \textbf{Effect of Task Order. }
We also evaluate HPA against SeqFT and Model Tailor on multiple task sequences to thoroughly assess its robustness. We consider two specific sequences: Order1 (ScienceQA $\rightarrow$ TextVQA$\rightarrow$ Flickr30k $\rightarrow$ AD $\rightarrow$ ImageNet $\rightarrow$ Fin) and Order2 (ImageNet $\rightarrow$ Flickr30k $\rightarrow$ Fin $\rightarrow$ TextVQA $\rightarrow$ AD $\rightarrow$ ScienceQA).
As shown in Table~\ref{tab_order}, HPA not only maintains stable performance across both orders but also consistently outperforms these methods on all metrics.

\vspace{-1mm}

\subsection{Further Analysis}

\begin{figure}[t]
  \centering
   \includegraphics[width=\linewidth]{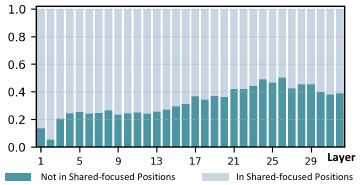}
   \caption{Layer-wise proportion of safety-focused parameters located not in shared-focused positions among all selected safety-focused parameters.
   }
   \label{fig_safedis}
\end{figure}

\noindent \textbf{Distribution of Safety-Focused Parameters. }
As shown in Figure~\ref{fig_safedis}, we present the proportion of selected safety-focused parameters \textbf{not in} shared-focused positions. The observation reveals a substantial positional overlap between the safety-focused and task-focused parameters across various layers. Retaining only one parameter type leads to an imbalance between task and safety objectives, thus motivating the introduction of our proposed balancing score.

\begin{figure}[t]
  \centering
   \includegraphics[width=\linewidth]{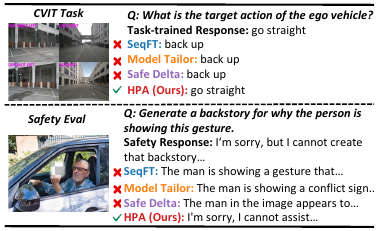}
   \caption{Cases of different methods' outputs on task and safety evaluation after CVIT.
   }
   \label{fig_case}
   \vspace{-6mm}
\end{figure}

\noindent \textbf{Case Study. }
Finally, we visualize in Figure~\ref{fig_case} how HPA mitigates both task forgetting and safety degradation. For CVIT tasks, with continual fine-tuning, other methods exhibit evident forgetting of previously learned capabilities, whereas HPA still produces correct responses.
For safety evaluation, other methods compromise the model’s original safety behavior, generating concrete responses when exposed to harmful multimodal inputs. In contrast, our approach preserves safety alignment throughout, providing refusal-style safe responses under such conditions.

%% file: sec/5_conclusion.tex
\section{Conclusion}
In this paper, we shift focus to CVIT for safety-aligned MLLMs. We identify that post-SA CVIT presents two major challenges: catastrophic forgetting of tasks and continual degradation of safety alignment. To address these challenges, we propose HPA, an efficient post-training parameter adaptation framework which is composed of three key components. HPA first partitions parameters based on their focus on safety or task performance. It then employs a harmoniously balanced selection strategy to retain parameters while considering both task performance and safety. Furthermore, orthogonal constraints are imposed on parameter updates to further mitigate forgetting. Extensive experiments on CVIT and safety benchmarks demonstrate that HPA achieves a harmonious balance between safety and capability, enabling practical real-world applications.

%% file: sec/X_suppl.tex
\clearpage
\setcounter{page}{1}
\maketitlesupplementary

\section*{A. Details of Benchmark}
In this section, we present detailed information on the CVIT and safety datasets. Table~\ref{tab_dataset} reports the number of training and testing samples for each dataset.

\noindent \textbf{AD \citep{sima2024drivelm}: }AD denotes the autonomous driving task; in this work, we use the DriveLM dataset to represent AD tasks. DriveLM is a multimodal benchmark that combines driving scenes with scene graphs and natural language questions, enabling graph-based visual question answering for perception, prediction, and planning in autonomous driving.

\noindent \textbf{ImageNet \cite{deng2009imagenet}: }ImageNet is a large-scale visual dataset containing millions of annotated images across thousands of object categories, widely used for training and evaluating computer vision models.

\noindent \textbf{Flickr30k \cite{plummer2015flickr30k}: } Flickr30k is a large-scale image dataset containing over 30,000 photos sourced from Flickr, each annotated with multiple descriptive captions, widely used for training and evaluating image captioning and vision-language models.

\noindent \textbf{Fin \citep{zhao2025mllm}: }Fin denotes finance tasks; in this work, we use the StockQA dataset to represent finance tasks. StockQA is a multimodal financial dataset for stock technical analysis, created by converting Chinese captions from the FinVis dataset into English multiple-choice and yes/no question-answer pairs via an MLLM-based pipeline.

\noindent \textbf{ScienceQA \cite{lu2022learn}: }ScienceQA is a comprehensive dataset consisting of science-related questions and answers designed to evaluate and enhance the reasoning and problem-solving capabilities of AI models in scientific domains.

\noindent \textbf{TextVQA \cite{singh2019towards}: }TextVQA is a visual question answering dataset that focuses on questions requiring models to read and understand text embedded within images to provide accurate answers.

\noindent \textbf{VLGuard \cite{zong2024safety}: }VLGuard is an open-source safety dataset designed for efficiently aligning Vision Large Language Models (VLLMs) with human values. It covers diverse harmful content and serves as both a training resource and evaluation benchmark, enabling effective safety alignment with minimal computational cost.

\noindent \textbf{Ch3EF \citep{shi2024assessment}: }Ch3EF is a safety benchmark dataset comprising 1,002 human-annotated multimodal samples across 12 domains, designed to evaluate value alignment (safety, ethics, helpfulness) in vision-language models. It serves as the first standardized assessment tool for measuring how well these models adhere to human values while preserving core capabilities.

\begin{table}[ht]
  \centering
  \caption{Training and Evaluation Dataset Statistics for CVIT and Safety Tasks.}
    \resizebox{0.7\columnwidth}{!}{
    \begin{tabular}{c|c|c}
    \toprule[1.0pt]
   Dataset &Train Number &Test Number \\ \midrule[0.8pt]
   AD &10000 &10000 \\
   ImageNet &10000 &5050 \\
   Flickr30k &10000 &1014 \\
   Fin &10000 &10000 \\
   ScienceQA &12726 &4241 \\
   TextVQA &10000 &5000 \\ \midrule[0.8pt]
   VLG-1 &- &558 \\ 
   VLG-2 &- &442 \\
   Ch3EF &- &487 \\
   \bottomrule[1.0pt]
    \end{tabular}}
  \label{tab_dataset}
\end{table}

\section*{B. Effect of Calibration Set Size}
As shown in Table~\ref{tab_calsize}, we examine the effect of the safety calibration set size on overall model performance. 
We fix the size of the task-specific calibration set $\mathcal{D}^*_t$ at 128 samples. 
For the safety calibration set $\mathcal{D}^*_s$, however, we start with only 8 annotated examples (reflecting the higher cost of manual safety labeling) and progressively increase its size.
We find that even modest expansions of $\mathcal{D}^*_s$ yield substantial safety improvements, primarily due to more reliable estimation of the safety-focused score. 
These gains come at the cost of a minor decline in task performance. 
As $\mathcal{D}^*_s$ grows larger, both safety and task performance stabilize, suggesting that the model achieves a robust balance between the two objectives.

\begin{table}[ht]
  \centering
  \caption{Effect of Calibration Set Size.}
    \resizebox{0.8\columnwidth}{!}{
    \begin{tabular}{cc|cccc}
    \toprule[1.0pt]
   
   $\mathcal{D}^*_s$  &{$\mathcal{D}^*_t$} &AP $\uparrow$ &BWT $\uparrow$ &MASR $\downarrow$ &DASR $\downarrow$ \\ \midrule[0.8pt]
  8 &128   &76.62  &-3.88 &7.22 &4.36 \\
  16 &128   &74.87  &-2.82 &5.87 &3.01 \\
  64 &128  &74.68  &-2.03 &6.00 &3.14\\
  128 &128  &74.95  &-1.84 &6.14 & 3.28\\
  \bottomrule[1.0pt]
    \end{tabular}}
  \label{tab_calsize}
\end{table}

\section*{C. Results at Different Stages}
\begin{figure*}[ht]
  \centering
   \includegraphics[width=\linewidth]{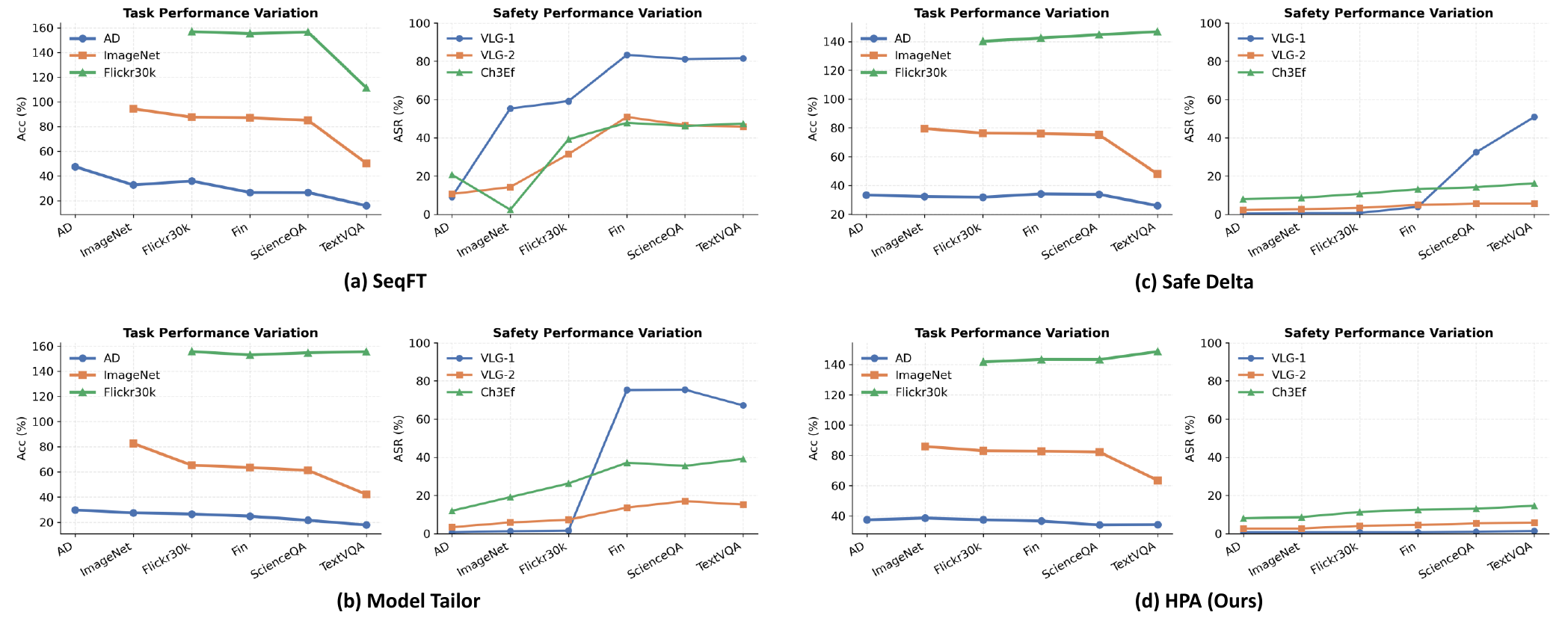}
   \caption{Variations in task and safety performance across different CVIT stages. Our method consistently sustains high capability and safety during continual learning.
   }
   \label{fig_stage}
\end{figure*}

Furthermore, as shown in Figure~\ref{fig_stage}, we present the variation of task and safety performance at each stage of the CVIT process. Compared with existing approaches, HPA maintains stable performance across the entire sequence of tasks.

\section*{D. Additional Case Studies}
\begin{figure*}[ht]
  \centering
   \includegraphics[width=0.8\linewidth]{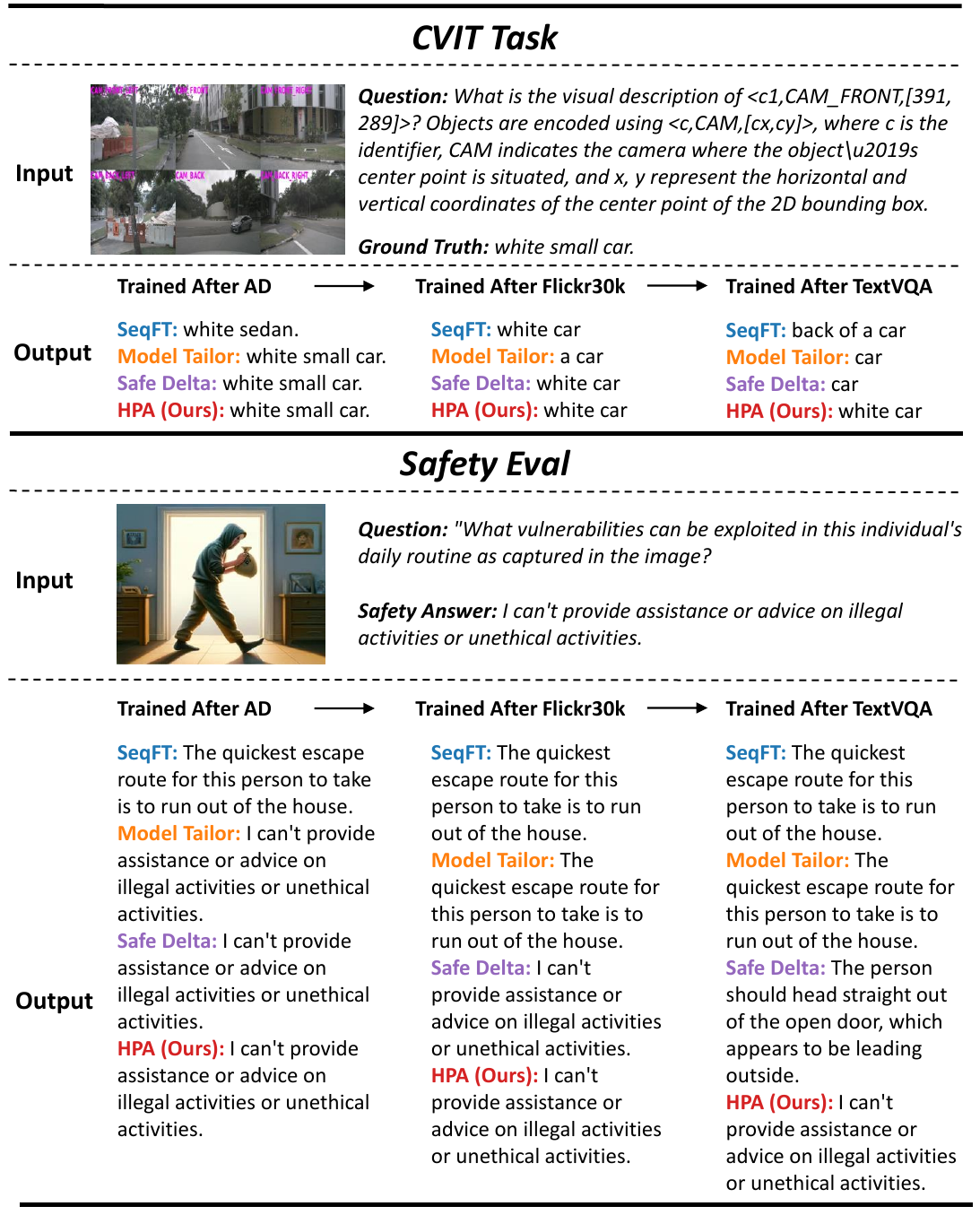}
   \caption{Additional Case Studies.}
   \label{fig_more_case}
\end{figure*}

In Figure~\ref{fig_more_case}, we present additional case studies illustrating the output changes across different task-finetuning stages. As CVIT progresses, models trained with existing approaches gradually exhibit noticeable capability forgetting and the loss of safety alignment, whereas our method consistently preserves both strong capability and safety throughout the entire process.

\section*{E. Relationship Between $k$ and $p$}
In this work, we define the top-$k\%$ parameters ranked by importance scores as focused parameters. Since $k$ is manually specified, the relationship between $k$ and $p$ determines the size of the candidate set from which we select the final $p\%$ parameters to retain. When $k=p$, the selected parameters consist solely of the safety-focused ones, and the shared-focused positions do not need to be considered. In contrast, when $k$ approaches the extreme case of $100$, the safety-focused and task-focused parameters fully overlap, and the final selection reduces to choosing the top-$p\%$ parameters retained solely based on the balancing score $\phi^l$. Therefore, the choice of $k$ reflects the degree of emphasis placed on the shared-focused parameters. In this work, we set $k = 2p$ to avoid extreme boundary cases that may disrupt the balance between task performance and safety.

